\documentclass[runningheads]{llncs}

\usepackage{iciap}

\usepackage{iciapabbrv}

\usepackage{graphicx}
\usepackage{booktabs}

\usepackage[accsupp]{axessibility}  

\usepackage{hyperref}

\usepackage{orcidlink}

\usepackage[T1]{fontenc}

\usepackage{times}
\usepackage{epsfig}
\usepackage{amsmath}
\usepackage{amssymb}
\usepackage{booktabs}
\usepackage[accsupp]{axessibility}
\usepackage{subcaption}
\usepackage{adjustbox}

\usepackage{graphicx}
\usepackage{txfonts}
\usepackage{hyperref}

\usepackage{color}
\usepackage{float}
\usepackage[rflt]{floatflt} 
\usepackage{mathtools}
\usepackage[normalem]{ulem}
\usepackage{float} 

\newcommand{\off}[2]{{}{}}

\usepackage[ruled,vlined,lined,linesnumbered]{algorithm2e}
\usepackage{dsfont}

\usepackage[capitalize]{cleveref}
\crefname{section}{Sec.}{Secs.}
\Crefname{section}{Section}{Sections}
\Crefname{table}{Table}{Tables}
\crefname{table}{Tab.}{Tabs.}

\begin{document}
\title{DeepCS-TRD, a Deep Learning-based Cross-Section Tree Ring Detector}
%
%
\author{
Henry Marichal\inst{1} \and
Verónica Casaravilla\inst{2} \and
Candice Power\inst{3}\and
Karolain Mello\inst{2} \and 
Joaquín~Mazarino\inst{2} \and 
Christine Lucas\inst{2} \and
Ludmila Profumo\inst{2} \and
Diego Passarella\inst{2} \and 
Gregory~Randall\inst{1}
}
\authorrunning{Marichal et al.}
%
\institute{
Facultad de Ingeniería, Universidad de la República,  Uruguay \and
CENUR, Universidad de la República, Uruguay\and
Ecoinformatics and Biodiversity, Department of Biology, Aarhus University, Denmark
}

\maketitle              
\begin{abstract}

Here, we propose Deep CS-TRD, a new automatic algorithm for detecting tree rings in whole cross-sections. It substitutes the edge detection step of CS-TRD 
by a deep-learning-based approach (U-Net), which allows the application of the method to different image domains: microscopy, scanner or smartphone acquired, and species (\textit{Pinus taeda}, \textit{Gleditsia triachantos} and \textit{Salix glauca}). Additionally, we introduce two publicly available datasets of annotated images to the community. 
The proposed method outperforms state-of-the-art approaches in macro images (\textit{Pinus taeda} and \textit{Gleditsia triacanthos}) while showing slightly lower performance in microscopy images of \textit{Salix glauca}. To our knowledge, this is the first paper that studies automatic tree ring detection for such different species and acquisition conditions. The dataset and source code are available in \url{https://hmarichal93.github.io/deepcstrd/}.

\end{abstract}

\section{Introduction}

Most existing automatic tree ring measurement methods rely on images from cores (small cylindrical samples that capture a transect of the tree's growth rings) rather than entire cross-sections. However, extracting a core is not feasible for very small trees or shrubs, and analyzing entire cross-sections is essential for individuals with suppressed or irregular growth, where there may be wedging rings. Furthermore, modeling tree growth in forestry often requires 2D information, which can only be obtained from analyzing the entire cross-section.

\begin{figure}[ht]
\begin{center}
   \begin{subfigure}{0.23\textwidth}
   \includegraphics[width=\textwidth]{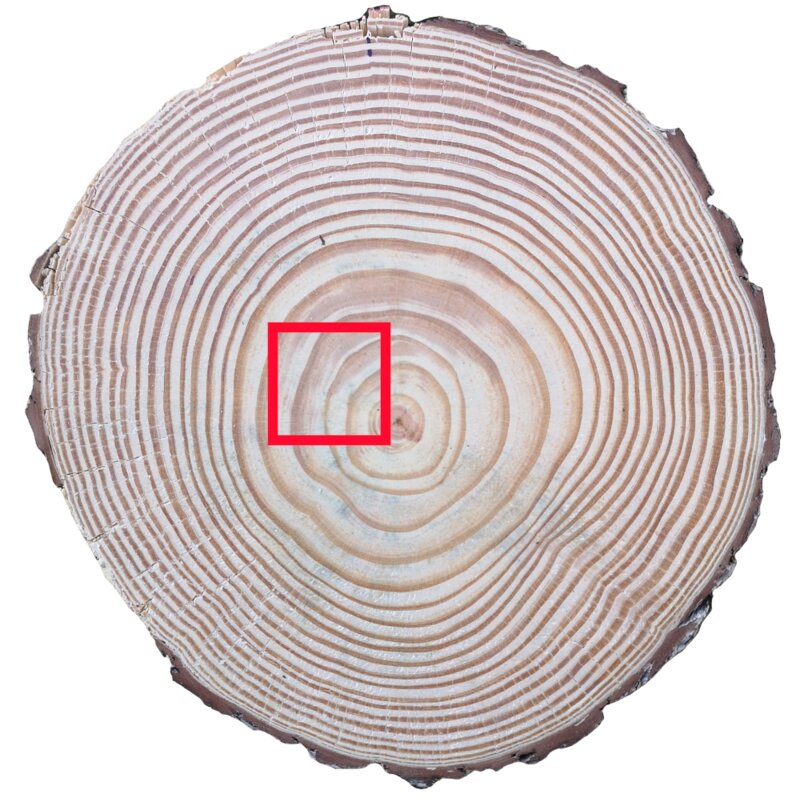}
   \caption{}
   \end{subfigure}
   \begin{subfigure}{0.23\textwidth}
   \includegraphics[width=\textwidth]{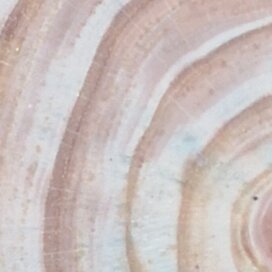}
   \caption{}
   \end{subfigure}
   \begin{subfigure}{0.23\textwidth}
   \includegraphics[width=\textwidth]{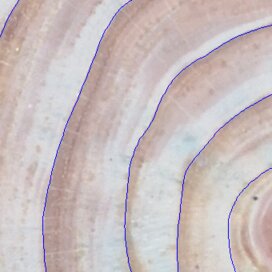}
   \caption{}
   \end{subfigure}

   \caption{\textbf{Tree Ring Detection}. Accurate detection of tree ring boundaries in images is critical. (a) Wood cross-section image. (b) Zoomed-in view of the red square in (a). (c) The same view with ring boundaries in blue. The ring thickness is set to 1 pixel.} 
   \label{fig:rings_details}
\end{center}
\end{figure}

Automatically delineating rings in images of tree cross-sections presents the challenge of generating a pattern of closed curves that accurately represent the ring boundaries, as seen in \Cref{fig:rings_details}. These rings may exhibit irregular and asymmetric growth patterns, especially in trees and shrubs growing in extreme environments. Moreover, ring boundaries make up only a small portion of the disc, which may also contain perturbations such as cracks, knots, and fungal growth, further increasing the complexity of the task. 

\begin{figure}[ht]
\includegraphics[width=\textwidth]{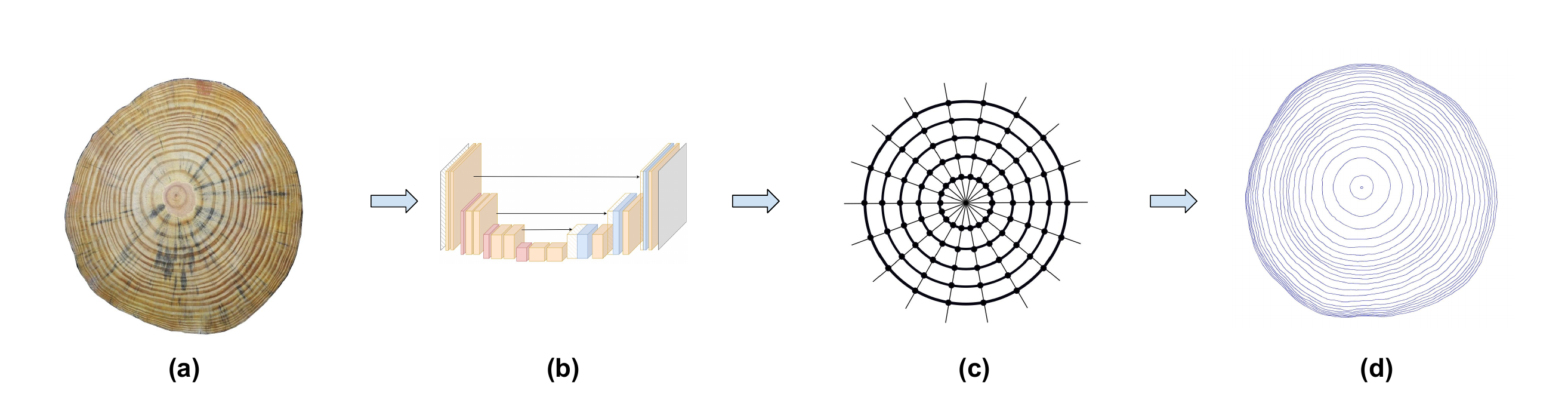}
\caption{\textbf{DeepCS-TRD Tree-Ring detection pipeline}. (a) Input disc image. (b) Deep Contour Detector. (c) \textit{Spider web} model. (d) Tree-ring continuous curves.} 
\label{fig:pipeline}
\end{figure}

\Cref{fig:pipeline} outlines the pipeline proposed in this work: a disc image is provided as input, and tree ring traces are generated as output. Once the tree ring curves are delineated, areas and perimeter can be computed, which is essential for forestry studies \cite{atmos14020319}. 
Marichal et al. \cite{marichal2023cstrd} proposed an automatic method for tree ring delineation in wood cross-section images called CS-TRD. This method is based on classical edge detection techniques and was applied to two coniferous tree species, \textit{Pinus taeda} and \textit{Abies alba}.  Furthermore, they introduced a model for wood cross-section images named \textit{spider web}, see \Cref{fig:pipeline}c. This model enforces constraints to ensure that tree rings do not intersect and remain roughly concentric.

In this work, the edge detection module from the CS-TRD method is replaced with a U-Net \cite{unet} model, a deep learning architecture designed explicitly for semantic segmentation, as illustrated in \Cref{fig:pipeline}b. U-Net architecture has demonstrated remarkable results in tree ring delineation of cores \cite{Fabija_ska_2018} and in microscopy images of shrubs \cite{inbd}.

\begin{figure}[ht]
\begin{center}
   \begin{subfigure}{0.19\textwidth}
   \includegraphics[width=\textwidth]{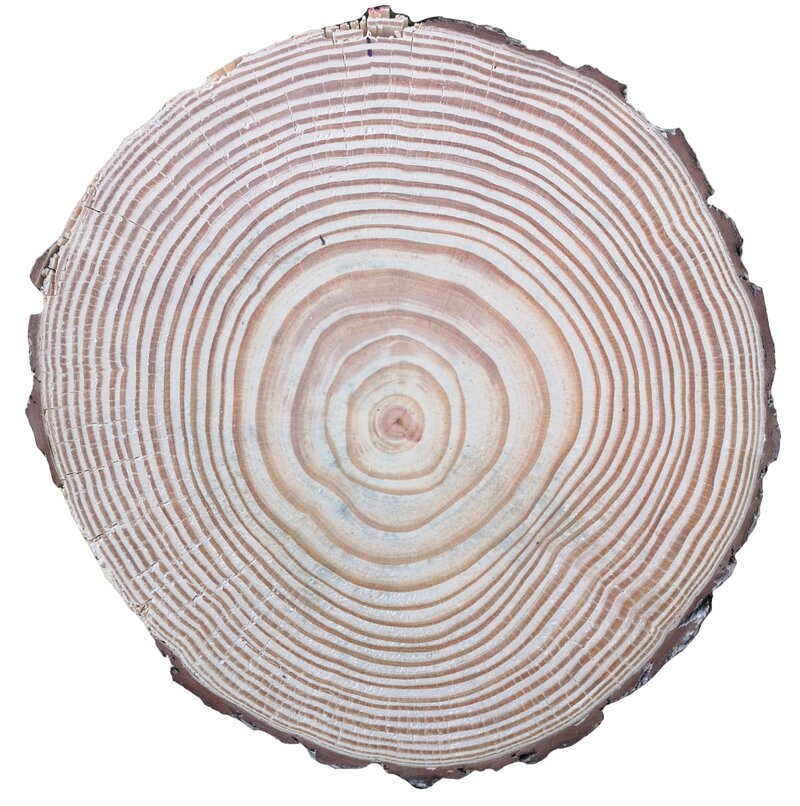}
   \caption{UruDendro1}
   \label{fig:ExForest}
   \end{subfigure}
   \begin{subfigure}{0.19\textwidth}
   \includegraphics[width=\textwidth]{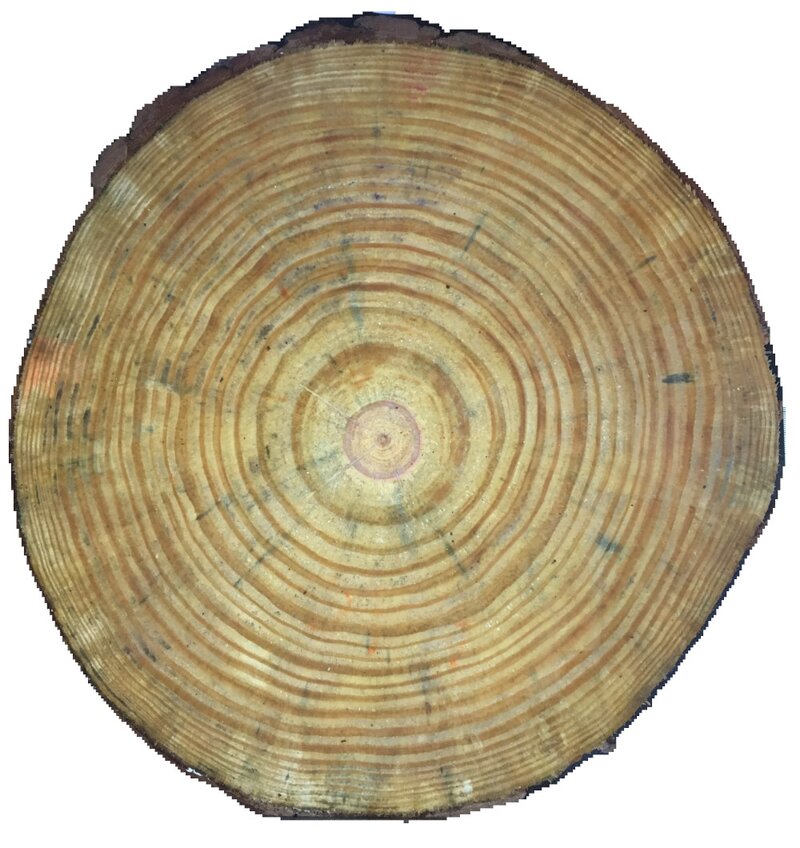}
   \caption{UruDendro2}
   \label{fig:ExLongyard}
   \end{subfigure}
    \begin{subfigure}{0.19\textwidth}
   \includegraphics[width=\textwidth]{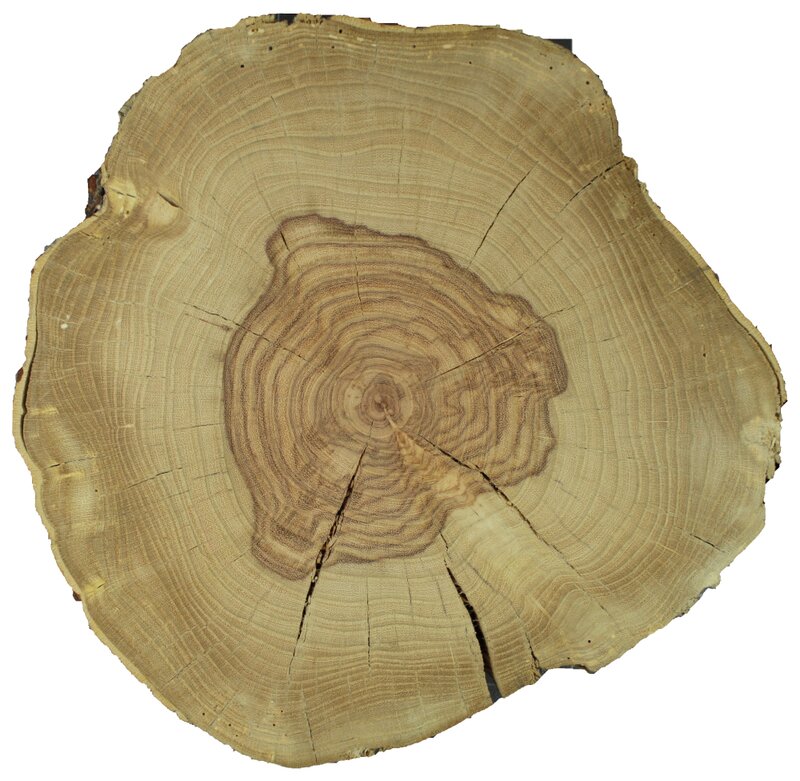}
   \caption{UruDendro3a}
   \label{fig:ExLogs}
   \end{subfigure}
   \begin{subfigure}{0.19\textwidth}
   \includegraphics[width=\textwidth]{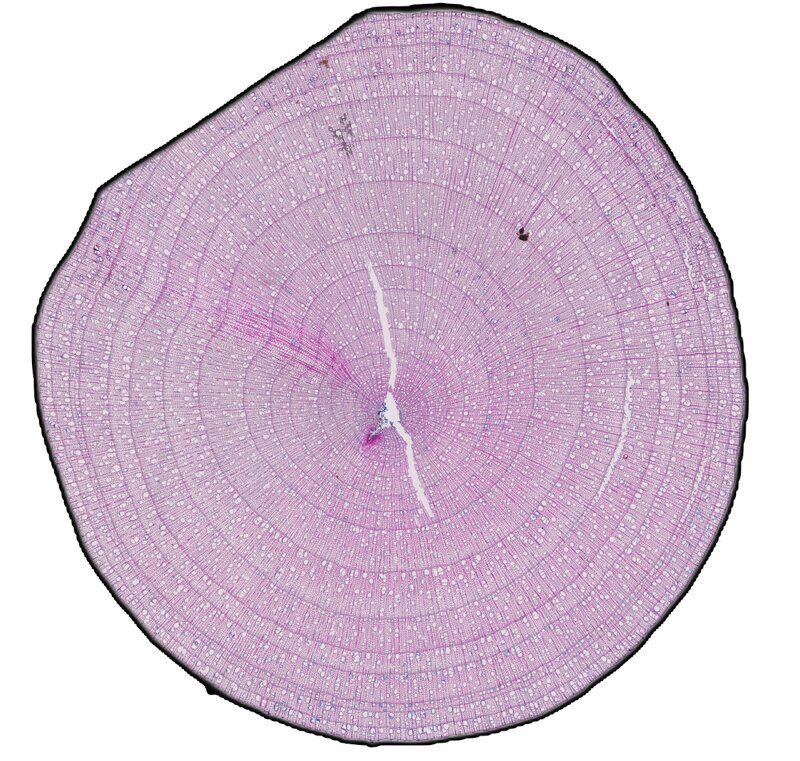}
   \caption{DiskoIsland}
   \label{fig:ExGleditsia}
   \end{subfigure}
   \hfill
   \caption{\textbf{Examples of the used datasets}. Species are (a-b) \textit{Pinus taeda}, (c) \textit{Gleditsia triacanthos}, (d) \textit{Salix glauca}. Acquisition conditions: (a-c) in a laboratory with a smartphone camera, sanded and polished. (d) microsection stained with 1 \% safranin and 0.5 \% astrablue, permanently fixed to a microscope slide with Eukitt, and scanned at 100 x magnification}. 
   \label{fig:bbdd}
\end{center}
\end{figure}

By replacing the edge detection module in the CS-TRD method with this deep learning-based model, we successfully extended its application to a diverse range of species and growth forms. Notably, this enhanced approach works with angiosperm trees (i.e. \textit{Gleditsia triacanthos}) and shrubs prepared using microscopy (i.e. \textit{Salix glauca})\\, samples illustrated in \Cref{fig:bbdd}. We refer to this new method as DeepCS-TRD.

In addition, we introduce a new image dataset of the \textit{Pinus taeda} species (UruDendro2 \cite{uru2}), containing 53 samples, and another dataset comprising nine samples of \textit{Gleditsia triacanthos} (UruDendro3a \cite{uru3a}), all
with their corresponding expert ring delineations. Generating these types of datasets is a highly time-consuming task that requires the expertise of dendrochronology specialists and is critical for algorithm training.

\section{Previous work}

Though not yet widely adopted in the tree-ring community, deep learning approaches have become more prevalent in recent years and have naturally been applied to this problem, framing ring detection tasks as a segmentation problem. Two architectures are commonly used: U-Net and Mask R-CNN \cite{unetvsmaskrcnn}. Gillert et al. \cite{inbd} proposed a method for cross-section tree-ring detection called Iterative Next Boundary Detection Network (INBD), based on the U-NET architecture. This method was applied to high-resolution microscopy images of shrub cross-sections, detecting the annual rings individually at each iteration step, from the medulla to the bark. INBD was trained and tested on shrub microscopy images, using standard cross-entropy loss as the main loss for the ring detection network.

Besides the INBD method, most deep-learning-based approaches have been applied to core images. Polek et al.\cite{Pol_ek_2023} used a Mask R-CNN deep learning architecture to detect rings in cores of coniferous species. During network training, they used mean average precision (mAP) as a loss function. Fabijańska et al. proposed a classic image-processing approach \cite{FABIJANSKA2017279} (based on the linking of maximum image gradient pixels) and a convolutional neural network based on a U-Net architecture \cite{Fabija_ska_2018} for detecting tree rings in core images. When comparing both methods, they reported a significant improvement in precision and recall for the deep learning approach over the classic one. The U-Net network was trained to minimize the categorical cross-entropy loss function.

The scarcity of annotated tree ring image datasets remains a significant challenge, as the methods must be tailored to the specific characteristics of each species. Gillert et al.\cite{inbd} made a dataset available of 213 high-resolution annotated images of cross-sections from three shrub species (\textit{Dryas octopetala}, \textit{Empetrum hermaphroditum}, and \textit{Vaccinium myrtillus}), that were prepared using similar methods to those applied to the \textit{Salix glauca} included in this study (see \Cref{fig:bbdd}d). Furthermore, the CS-TRD publication included a publicly available dataset of 64 annotated images of \textit{Pinus taeda} (see \Cref{fig:bbdd}a). Kennel et al. \cite{KENNEL2015204} included a dataset of 7 cross-section images of \textit{Abies alba}, although the ring annotations are not fully accessible. Regarding cores, Polek et al. \cite{Pol_ek_2023} made available a dataset of 2601 image patches (similar to the red square in \Cref{fig:rings_details}a) with ring annotations for \textit{Picea abies}, a coniferous tree species.

Among the available methods, only INBD and CS-TRD are adapted for processing full cross-section images and provide accessible code, allowing for comparison with our proposed approach.

\section{Approach}

\begin{algorithm}[ht]
    \caption{Deep Contour Detector}
    \label{algo:DLRingSegmentation}
    \KwIn{
        $Im_{in}$: RGB disc image\;
        $tile\_size$: Tile size (0 if no patching is applied)\;
        $total\_rotations$: Number of rotations applied to the image\;
        $cy$, $cx$: x and y coordinates of the pith position\;
    }
    \KwOut{
        $l\_ch_s$: list of continuous curve edges\;
        $l\_nodes_s$: list of nodes of the curves;

    }
        
    $\Omega \leftarrow \text{arange}(0, 360, 360 / total\_rotations)$; // Define rotation angles
    
    $P \leftarrow \text{Zero matrix of the same size as } Im_{in}$; // Initialize the probability map
    
    \tcp{Process each rotation angle}
    \For{$\theta \in \Omega$}{
        $I_{rotated} \leftarrow \text{rotateImage}(Im_{in}, cy, cx, \theta)$; // Rotate the image around the pith position
        
        $P_{rotated} \leftarrow \text{ringSegmentationModel}(I_{rotated}, tile\_size)$; // Apply the segmentation model. See \Cref{fig:inference}.
        
        $P_{aux} \leftarrow \text{rotateImage}(P_{rotated}, cy, cx, -\theta)$; // Undo the rotation of the probability map
        
        $P \leftarrow P + P_{aux}$; // Accumulate the rotated probability map
    }
    $P \leftarrow P / total\_rotations$; // Average the probability map
    
    $M \leftarrow P  \geq 0.2$; // Threshold the accumulated probability map
    
    $S$ $\leftarrow \text{skeletonize}(M)$; // Skeletonize the binary mask
    
    $m\_ch_e \leftarrow \text{findContoursCurves}(S)$; // Extract ring curves from the skeleton
    
    $Nx, Ny \leftarrow \text{getNormalComponent}(m\_ch_e)$; // Compute normal directions for each edge

     $m\_ch_f \leftarrow \text{filterEdges}(m\_ch_e, Nx, Ny)$; // Filter edge curves. See \Cref{equ:condition}

     $l\_ch_s, l\_nodes_s \leftarrow$ samplingCurves($m\_ch_f$); // Convert curves to the \textit{spider web} format
    
    \KwRet{$l\_ch_s, l\_nodes_s$}; // Return the results
\end{algorithm}

The proposed approach aims to extract a set of continuous, connected pixel curves from an RGB image of a wood cross-section, a disc. These curves serve as input to the \textit{spider web} model presented in \cite{marichal2023cstrd}, enabling the reconstruction of digital tree-ring curves for the sample. An overview of the whole Tree-Ring detection pipeline is in \Cref{fig:pipeline}.

The CS-TRD method \cite{marichal2023cstrd} works as follows: (i) Several rays are traced departing from the center (the pith) outwards. (ii) Canny edge filtering is applied to the image. (iii) Filter out edges not belonging to the latewood to earlywood transitions, which correspond to the annual rings, using the angle between the edge normals and the rays.  (iv) All edge chains that belong to the same ring are grouped by imposing the \textit{spider web} structure through an iterative procedure that connects chains that do not cross each other, have strong edge information, and comply with a smoothness condition. The authors claim an F-Score of 89\% in \textit{Pinus taeda} and 97\% in \textit{Abies alba} species.

\subsection{Algorithm}

\begin{figure}[ht]
\centering 
\includegraphics[width=0.9\textwidth]{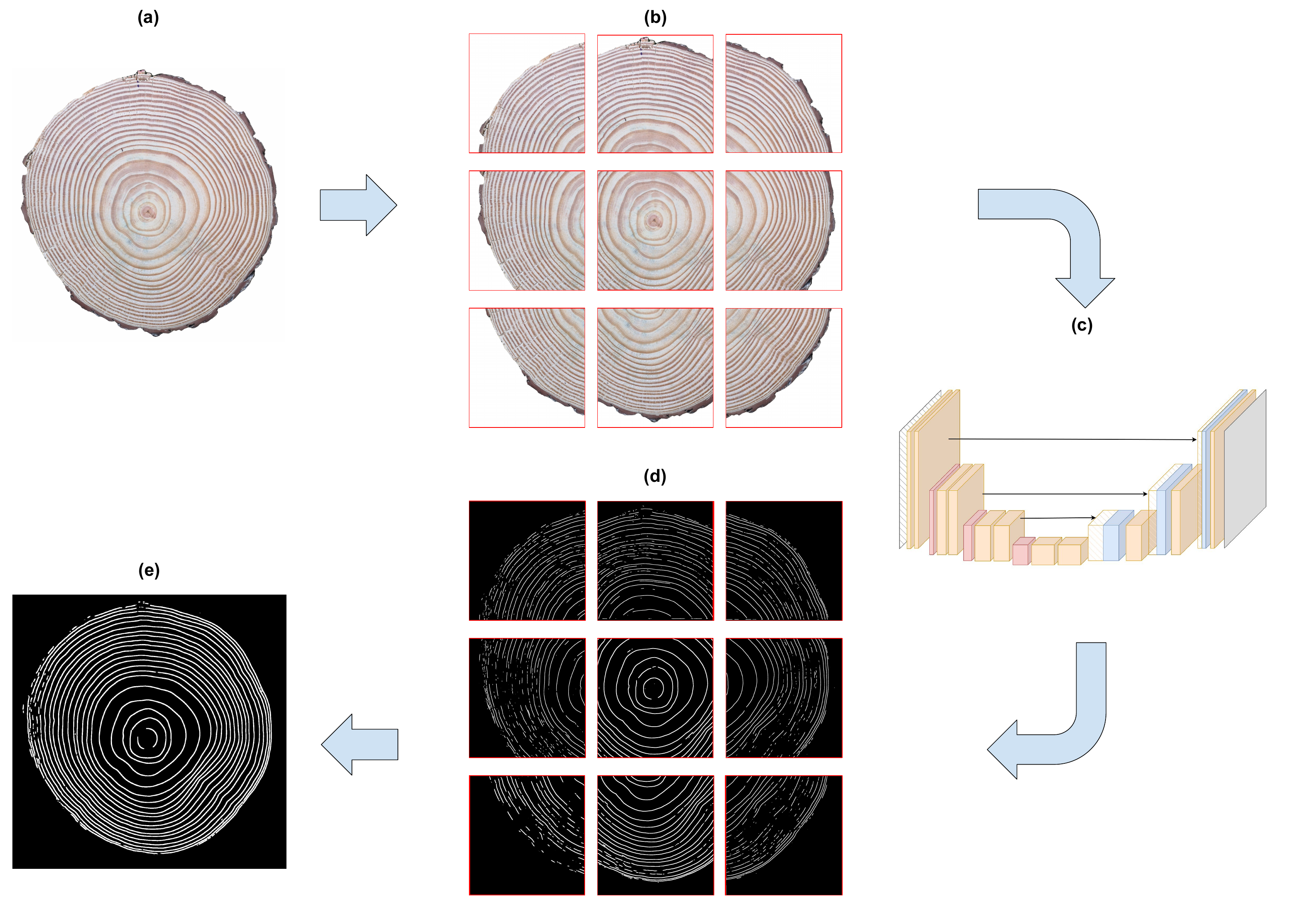}
\caption{\textbf{Ring Segmentation Model Inference.} The input to the model is a complete cross-section disc (a). (b) The disc is divided into square tiles. (c) and (d) Each tile is processed individually using the U-Net network to generate its corresponding probability map. 
Finally, the tile predictions are combined to reconstruct the probability map 
for the entire sample (e). This procedure corresponds to line 5 of \Cref{algo:DLRingSegmentation}.} 
\label{fig:inference}
\end{figure}

The DeepCS-TRD method maintains the principal steps of the CS-TRD 
 but with a substantial modification: replacing the Canny-based edge detection step with a module based on the U-Net architecture \cite{unet} trained to segment ring boundaries (\Cref{fig:pipeline}b). 
\Cref{algo:DLRingSegmentation} shows the pseudocode of this step. The method's inputs are an RGB image of a disc ($Im_{in}$) and a parameter to indicate the image is split into tiles of ($tile\_size$) size, following the overlap-tile strategy described in \cite{unet}. Additionally, the method allows to apply a Test-Time Augmentations (TTA) \cite{Kimura_2021} strategy, which consists of $total\_rotation$ rotations around the pith location ($c_y$, $c_x$).
The method produces a list of continuous ring edge curves ($l\_ch_s$) and a list of pixels nodes ($l\_nodes_s$) which are the input to the \textit{spyder web} model.
In lines 1 and 2, the method initializes the TTA rotation domain angles  $\Omega$ and the probability map $P$. Then, between lines 3 and 7, the inference loop iterates between the elements $\theta \in \Omega$. In line 4, $Im_{in}$ is rotated an angle $\theta$ and stored in $I_{rotated}$. In line 5, the ring boundaries are generated with the \textit{ring segmentation model} as illustrated in \Cref{fig:inference}.
First, if needed, the image is split into tiles with an overlapping of 10\% (to avoid the tile border effects). The tiles are zero-padded when necessary to ensure compatibility with the stride size ($tile\_size$). Then, a probability map is computed for each tile with a U-Net network. Finally, the probability map for the whole disc ($P_{rotated}$) is built (with the original image size).  The average between the tile's probability map is assigned to the overlapping area.

\begin{figure}[ht]
    \centering
    \begin{subfigure}[b]{0.3\textwidth}
        \centering
      \includegraphics[width=\textwidth]{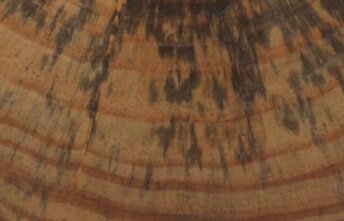}
        \caption{}
        \label{fig:patch_disk}
    \end{subfigure}
    \begin{subfigure}[b]{0.3\textwidth}
        \centering
       \includegraphics[width=\textwidth]{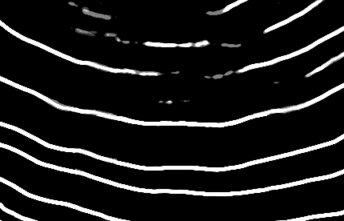}
        \caption{}
        \label{fig:patch_disk_map}
    \end{subfigure}
    
    \begin{subfigure}[b]{0.3\textwidth}
        \centering
       \includegraphics[width=\textwidth]{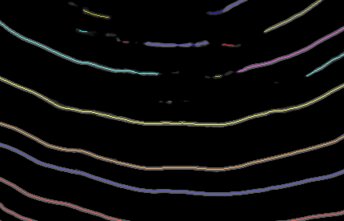}
        \caption{}
        \label{fig:patch_disk_skel}
    \end{subfigure}
    \begin{subfigure}[b]{0.3\textwidth}
        \centering
       \includegraphics[width=\textwidth]{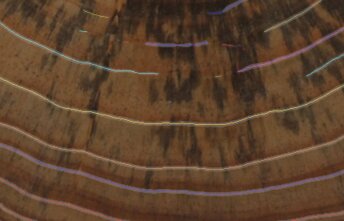}
        \caption{}
        \label{fig:patch_disk_skel_img}
    \end{subfigure}
    
    \caption{\textbf{Stages for extracting the ring edge curves.} (a) Zoomed-in view of the RGB image. (b) The probability map of the image section is shown in (a). (c) Probability map with overlaid edge curves. (d) Extracted edge curves overlaid on the image. }
    \label{fig:disk_posprocessing}
\end{figure}

Once the probability map of the whole image ($P_{rotated}$) has been obtained, the rotation of angle $\theta$ needs to be inverted (line 6, angle -$\theta$). Then, in line 7, the current probability map ($P_{aux}$) is accumulated into $P$. In line 8, the average probability map is obtained. 
In line 9, the probability map $P$ is binarized with a 0.2 threshold to generate the mask $M$. This low threshold is deliberately chosen, considering an additional filtering step in line 13 that further refines the results.
In line 10, this mask is skeletonized to improve the tree-ring curve precision. Finally, curves are extracted from the skeleton $S$ using the Suzuki method \cite{SUZUKI198532} implemented in the function \textit{findCountour} of the OpenCV library \cite{opencv_library}. 
\Cref{fig:disk_posprocessing} illustrates the lines 8 to 11 steps. \Cref{fig:disk_posprocessing}b shows the probability map $P$ for a disc region. Brighter pixels indicate a higher probability of belonging to a ring. \Cref{fig:disk_posprocessing}c shows the skeleton $S$ overlaid on the mask $M$, with colors representing curve labels. The skeleton corresponds to the center of the mask. Finally, \Cref{fig:disk_posprocessing}d presents the extracted curves overlaid on the original image.

The normal component of each pixel belonging to a curve in $m\_ch_e$ is computed using a straightforward procedure (line 12). The pixels of each curve are ordered sequentially from one end to the other. The tangent direction of pixel $p_0$ is
$\vec{T_{p_0}} = p_1 - p_{-1}$, where $p_{-1}$ and $p_1$ are the preceding and succeeding pixels along the curve, respectively. The normal direction $\vec{N_{p_0}}$ is then computed as a vector perpendicular to $\vec{T_{p_0}}$, given by $\vec{N_{p_0}} = \left(-T_{p_0,y}, T_{p_0,x}\right)$, where $\vec{T_{p_0}} = (T_{p_0,x}, T_{p_0,y})$. This ensures that $\vec{N_{p_0}}$ is orthogonal to the curve at $p_0$. The point $c$ is the pith location ($c_x,c_y$). In line 13, the angle $\delta$ between $\vec{cp_0}$ and $\vec{N_{p_0}}$ is computed as 
$ \delta\left(\vec{cp_0}, \vec{N_{p_0}}\right) = \arccos\left( \frac{\vec{cp_0} \cdot \vec{N_{p_0}}}{\|\vec{cp_0}\| \|\vec{N_{p_0}}\|} \right)\label{equ:filter_angle}   
$

We filter out all \textit{pixels} $p_0$ with normals not collinear and outbound oriented concerning the ray's direction at that point:

\begin{equation}
\alpha \leq \delta(\vec{c p_0},\vec{N_{p_0}})   \leq 180-\alpha
\label{equ:condition}
\end{equation}

We set $\alpha = 45$ degrees. 
When a pixel $p_0$ is removed, the curve to which $p_0$ belongs is split at that position. This step is straightforward because $m\_ch_e$ is a matrix of dimensions $N \times 2$. Each curve is separated from the others by a row with values $(-1,-1)$.
Therefore, if a pixel does not satisfy the condition in \Cref{equ:condition}, we insert the vector $(-1,-1)$  at its position.

Finally, in line 14, the curves in $m\_ch_f$ are converted into the format required by the \textit{spider web} model, structured around a central \textit{pith}, the origin of the \textit{rays}.  Each \textit{curve} is represented as a set of points, with \textit{nodes} in the intersections of the curve with the \textit{rays}. The resulting sampled \textit{curve} is called a \textit{chain}. These \textit{chains} have the notable property of being non-intersecting, inspired by the biological characteristics of tree rings observed in conifer species. More details can be found in \cite{marichal2023cstrd}.

\paragraph{Preprocessing} 
The background was removed from all images using the $U^2$-Net network \cite{QIN2020107404} and manually corrected when necessary. Furthermore, portions of the image background were cropped to focus on the cross-section so that the minimum distance between the disc and the image edges is 50 pixels. Images from all datasets were resized to 1504x1504 pixels using Lanczos interpolation. To complete the image size, if the smallest dimension of the resized image does not reach the new dimension, a 255-value padding was applied, maintaining the aspect ratio.

\subsection{Network Training}

Non-overlapping square tiles were used for training the U-Net network. Only the patches with ring presence of each image sample were used. The tile size is a hyperparameter of the training step (see \Cref{sec:experiments})  
where 60\% of the samples were used for training, 20\% for validation and 20\% for assessing the full tree-ring detection pipeline (\Cref{fig:pipeline}). Training and validation subsets were split into patches to train and select the network's hyperparameters. We also trained the network to predict the ring boundaries on the complete disc sample. 

The network was trained 
over 100 epochs on an NVIDIA P100 GPU with 12 GB of VRAM and 40 GB of system RAM. The Adam optimizer 
was used with an initial learning rate of 1e-3, while a cosine annealing schedule 
was applied to adjust the learning rate dynamically during training. The Dice Loss function was employed to optimize the model's performance. The Resnet18 backbone, pre-trained on ImageNet, was also used to enhance feature extraction. The network's weights that result in the lowest validation loss during training are selected as the optimal model.

Regarding data augmentation, in 50\% of the dataset images, randomly selected, we generate three augmented images: one with a random rotation around the pith, another with occlusions simulating cracks or fungi, and a third with elastic deformations.

\section{Datasets}

The datasets used in this study encompass a variety of species as shown in \Cref{fig:bbdd}. Some species, like the conifer \textit{Pinus taeda}, have clear ring contrasts, while  \textit{Gleditsia triacanthos} and \textit{Salix glauca} are more challenging. Due to the differences in growth forms, the sample preparation methods also varied. While the cross sections from the trees in datasets UruDendro 1, 2, and 3, were sanded and polished as typical for tree samples, the cross-sections in the DiskoIsland dataset were prepared using standard protocols for shrubs. 
Specifically, 12-15 $\mu$m thick cross-sections were cut, stained with 1\% safranin and 0.5\% astrablue, and permanently fixed to microscope slides using Eukitt (BiOptica, Milan). The image acquisition methods also varied. The images from the UroDendro 1, 2, and 3 datasets were captured with smartphone cameras (iPhone 6s and Huawei P20 Pro), while the images of the microsections from the DiskoIsland dataset were captured at 100x magnification using a Zeiss slide scanner (Axio Scan Z1, Zeiss, Germany (2.26 pixel/$\mu$m)).     

Ring boundaries in the selected acquired images were annotated using the Labelme Tool \cite{wada2024labelme}, which allows users to mark polylines over the image and export the annotations in JSON format. Ring annotations were performed by users with different levels of expertise, from undergraduate students to expert professors, and in all cases, they were reviewed by at least one other expert. Finally, a binary image mask was generated from the ring boundary annotations, with a thickness of 3 pixels.
\Cref{tab:patches} depicts the datasets, the number of samples and rings in each one, the species, and the acquisition method. 


\begin{table}
\centering
\caption{Dataset description. }\label{tab:patches}
\begin{tabular}{|l|c|c|c|c|}
\hline
\multicolumn{1}{|c|}{\textbf{Collection}} & \textbf{Samples} & \textbf{Tree-rings}  & \textbf{Species}& \textbf{Acquisition}\\ \hline
UruDendro1  \cite{uru1}                                & 64               & 1221                                 & \textit{Pinus taeda}           & Smartphone            \\ \hline
UruDendro2 \cite{uru2}                               & 53               & 1151                                      & \textit{Pinus taeda}           & Smartphone            \\ \hline
UruDendro3a \cite{uru3a}                               & 9                & 216                                      & \textit{Gleditsia triacanthos} & Smartphone            \\ \hline
DiskoIsland \cite{disko}                            & 50               & 654                                      & \textit{Salix glauca}          & Scanner              \\ \hline
\end{tabular}
\end{table}

\section{Results and discussion}

\subsection{Metrics}

We adopt the same evaluation metrics used by the authors of the INBD method: the mean Average Recall (mAR) and the Adapted Rand error (ARAND). False positives are preferred to false negatives because an incorrect ring can be deleted with just one click, while tracing a new one is more time-consuming. In this sense, the mAR metric is particularly suitable for this application.



\subsection{Experiments}
\label{sec:experiments}
The DeepCS-TRD method involved two key parameters:  $tile\_size$ and \textit{total\_rotations} (see \Cref{algo:DLRingSegmentation}).  We tested tile sizes of 64, 128, 256, 512, and 1504 (full resolution, no tiles) pixels for all the datasets. \Cref{fig:rings_details}a illustrates a tile size of 256px. 
For the \textit{total\_rotations} parameter, combinations of 0, 3, and 5 rotations were evaluated.  The selection of this parameter was made by optimizing the validation subset. After experiments, parameter values were fixed in $tile\_size=256$ and $total\_rotations=5$ for the four datasets.

We noted that the thickness of the annotated ring boundary masks during training impacts the method's performance. The best value was 3-pixel in all datasets. In consequence, DeepCS-TRD and INBD methods are trained using ring boundary thicknesses of three pixels. The two networks in the INBD method were trained following the author's procedure during 100 epochs, setting the downsampling parameter to 1, which gave the best results.

\begin{table}
\centering
\caption{\textbf{Results} on the test set for the four used datasets and the available SOTA methods: INBD \cite{inbd} and CS-TRD \cite{marichal2023cstrd} 
as well as the proposed DeepCS-TRD. 
The best performance is in boldface. Uru1 stands for UruDendro1, and the same is done for UruDendro2 and UruDendro3a datasets. Disko stands for DiskIsland dataset.}
\label{tab:results}
\begin{tabular}{lccccccccc}
 & \multicolumn{4}{c|}{\textbf{mAR} $\uparrow$}                               & \multicolumn{4}{c}{\textbf{ARAND} $\downarrow$}                             \\ \cline{1-9}
\multicolumn{1}{c|}{\textbf{Method}} & \textbf{Uru1} & \textbf{Uru2} & \textbf{Uru3a} & \multicolumn{1}{c|}{\textbf{Disko}} & \textbf{Uru1} & \textbf{Uru2} & \textbf{Uru3a} & \textbf{Disko} \\ \cline{1-9}
\multicolumn{1}{l|}{CS-TRD}          & .787          & .710                        & .007             & \multicolumn{1}{c|}{.026}             &      .093         &        .144                     & .466             & .634             \\ 
\multicolumn{1}{l|}{INBD}            &     .846          &  .742                           &          .200    &        \multicolumn{1}{c|}{\textbf{.735}}      &          .081     &          .132    &         .494      &             \textbf{.099}    &             \\ 

\multicolumn{1}{l|}{DeepCS-TRD}     & \textbf{.884}          & \textbf{.809 }                        &              \textbf{.620} & \multicolumn{1}{c|}{ .628 }         & \textbf{.053}          & \textbf{.105 }                       &  \textbf{.207 }            & .107         \\ 
\end{tabular}
\end{table}

\begin{figure}
\begin{center}
   \begin{subfigure}{0.3\textwidth}
   \includegraphics[width=\textwidth]{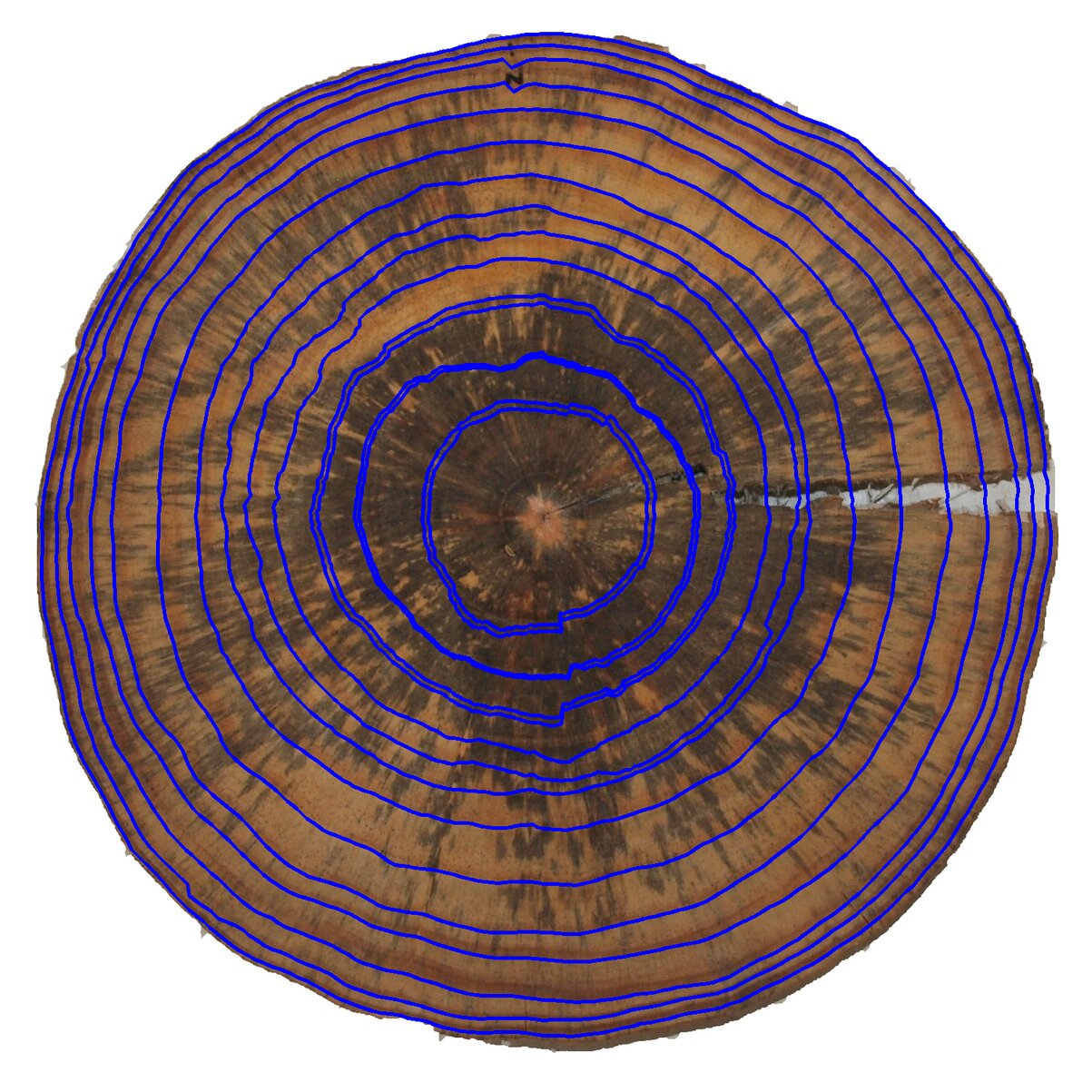}
   \end{subfigure}
   \begin{subfigure}{0.3\textwidth}
   \includegraphics[width=\textwidth]{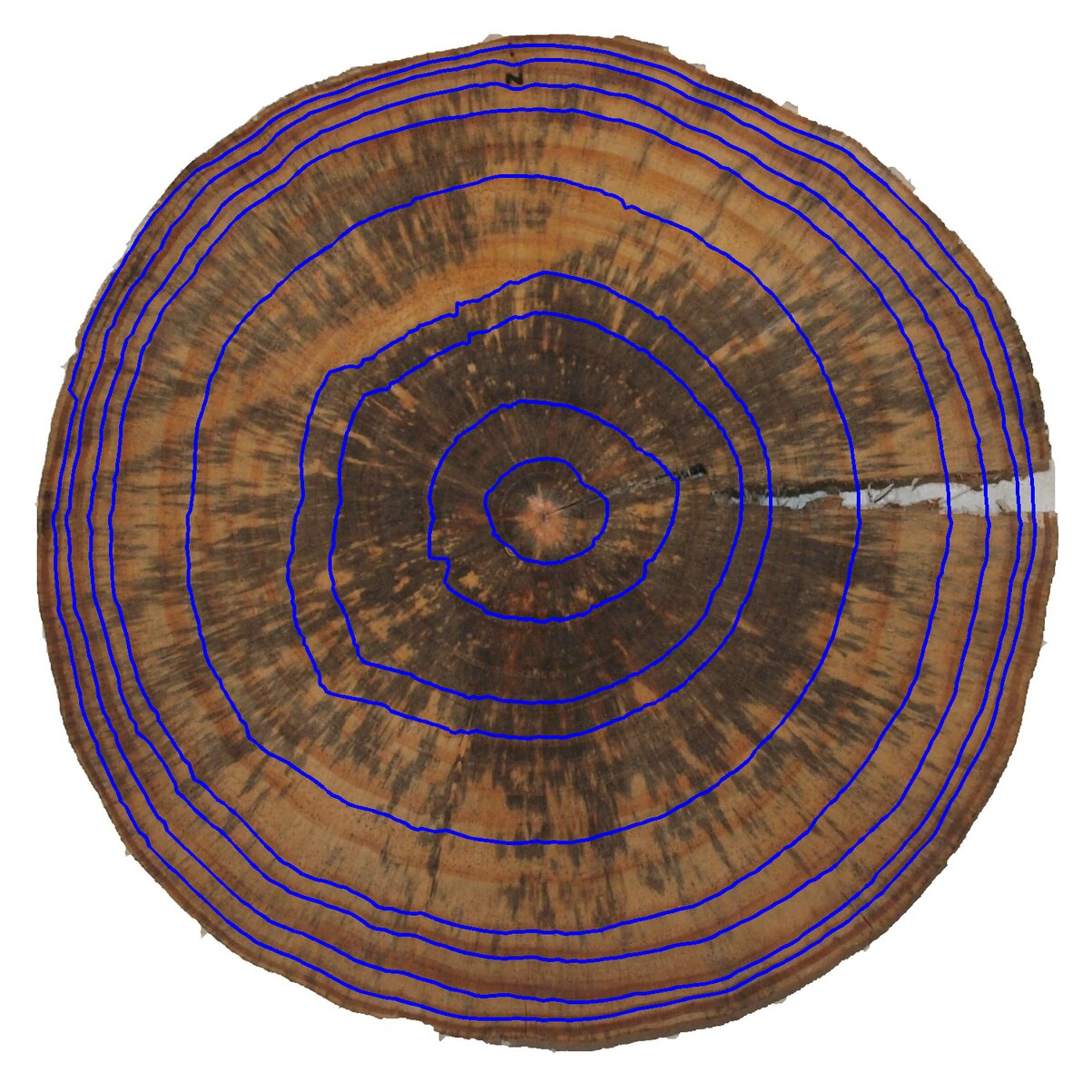}
   \end{subfigure}
   \begin{subfigure}{0.3\textwidth}
   \includegraphics[width=\textwidth]{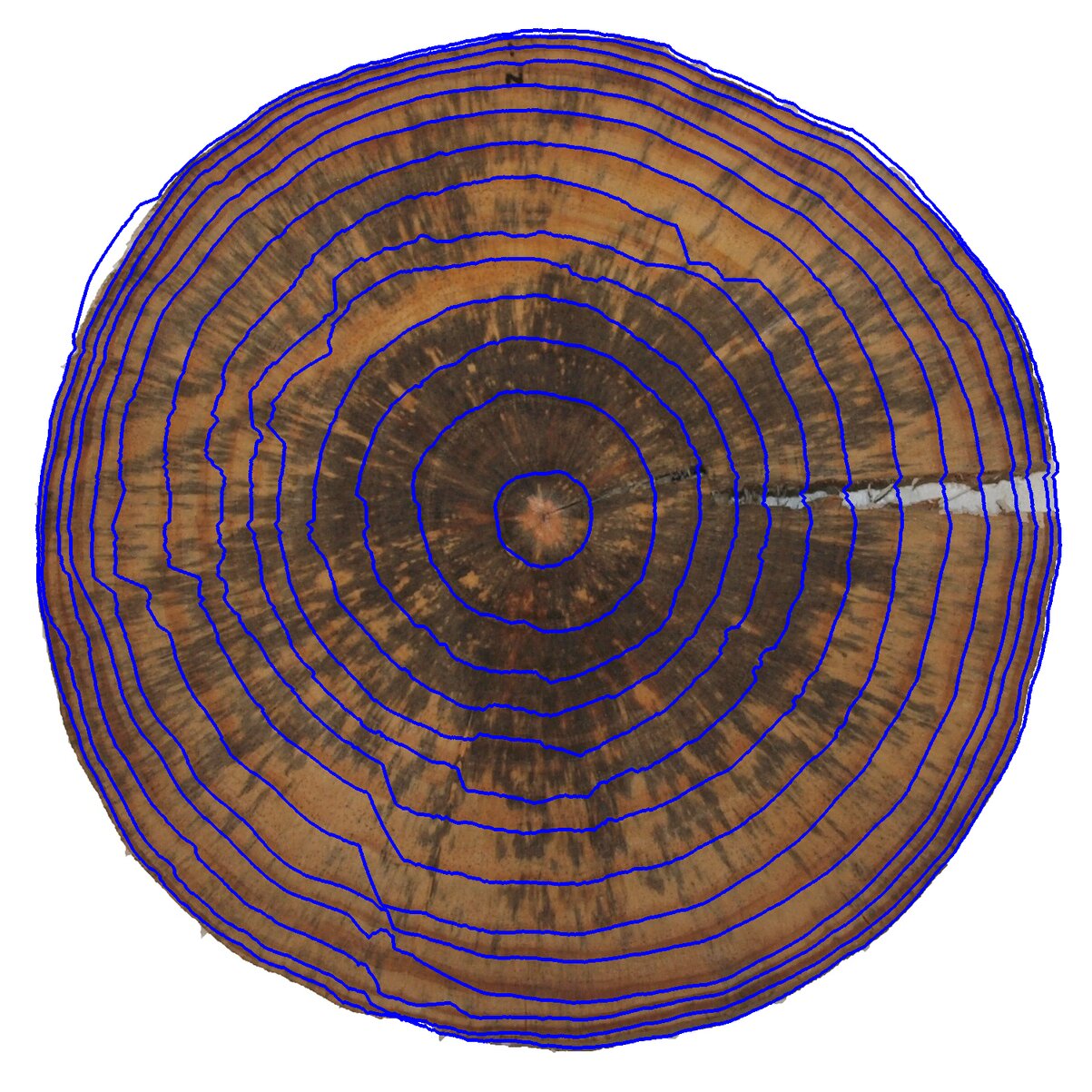}
   \end{subfigure}


    \begin{subfigure}{0.3\textwidth}
   \includegraphics[width=\textwidth]{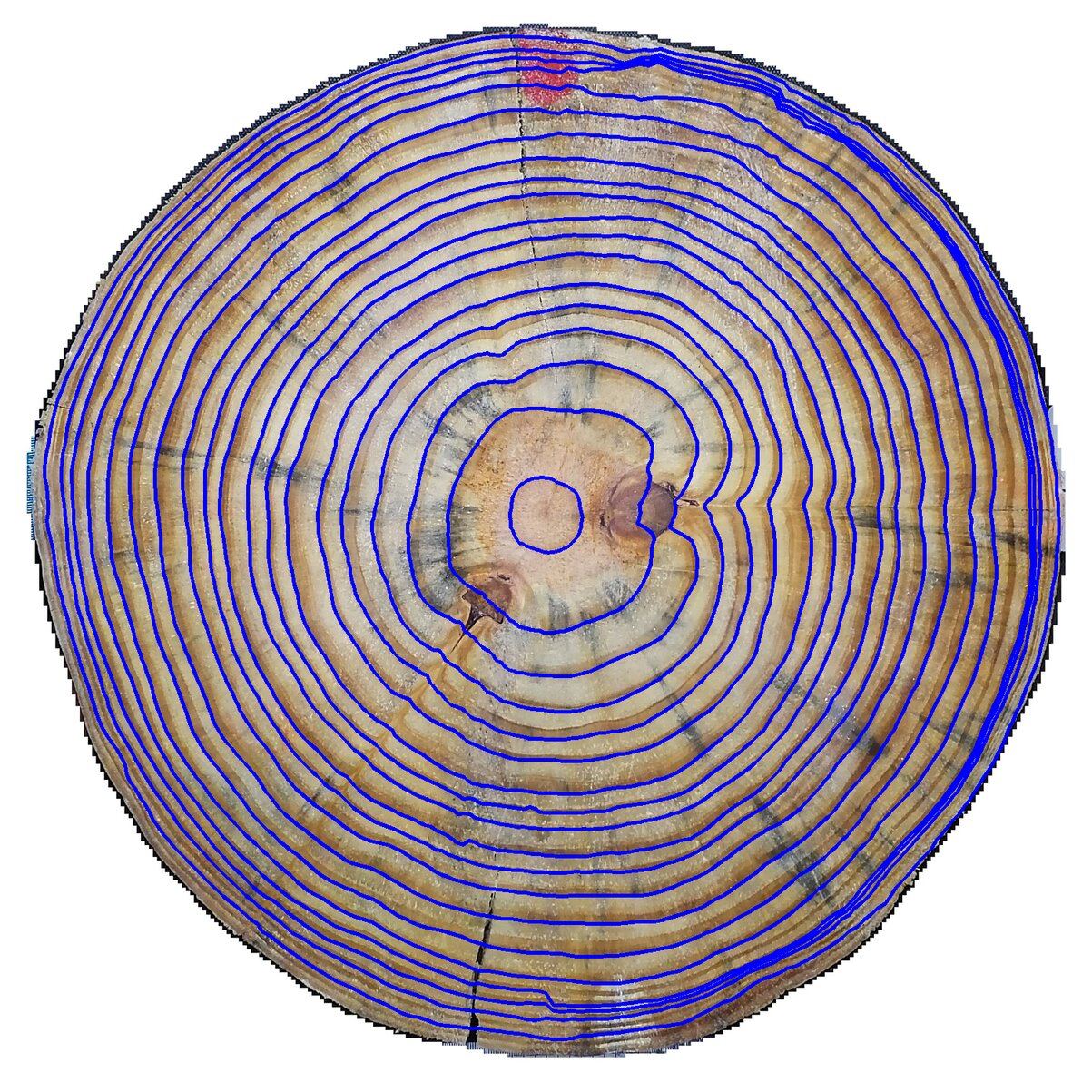}
   \end{subfigure}
   \begin{subfigure}{0.3\textwidth}
   \includegraphics[width=\textwidth]{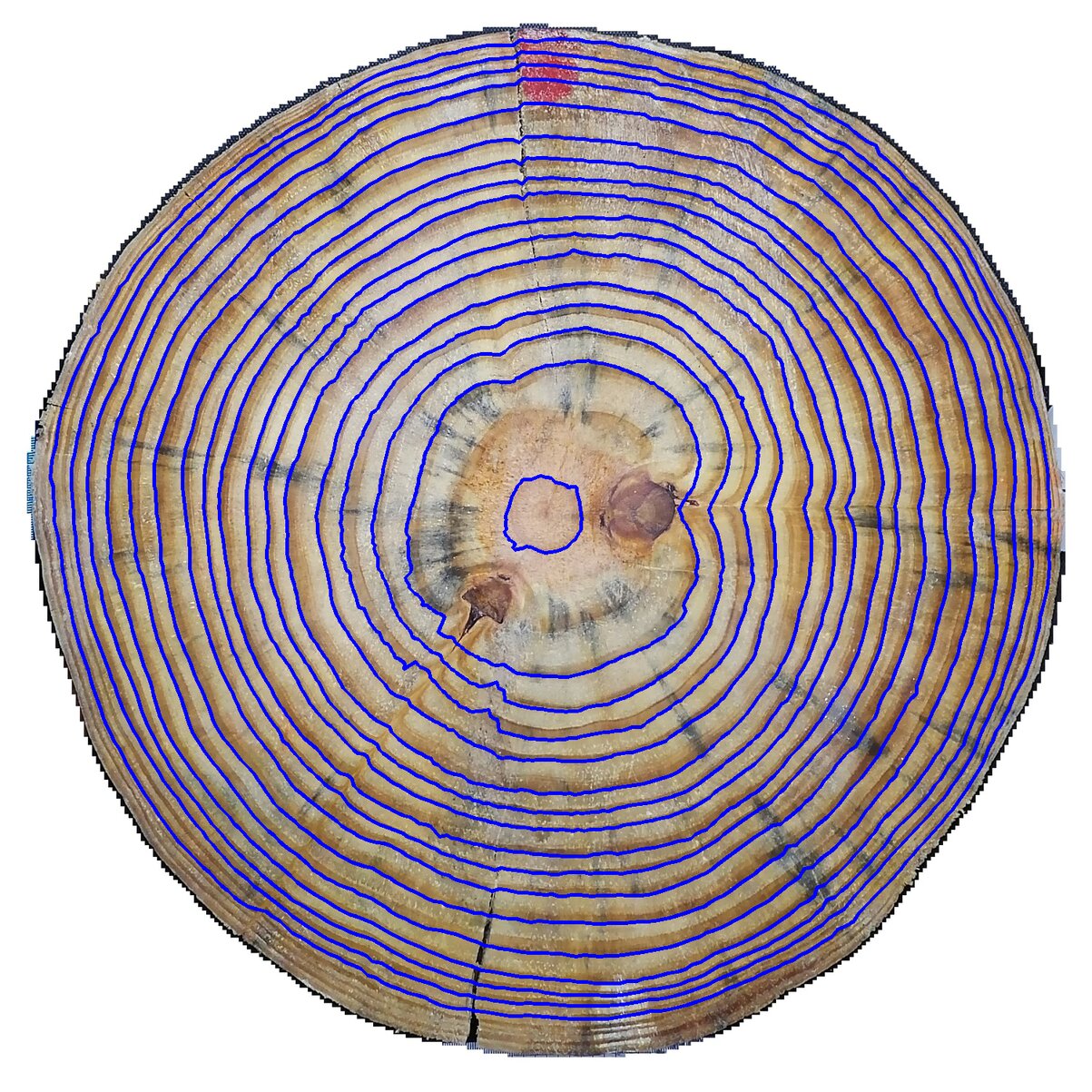}
   \end{subfigure}
   \begin{subfigure}{0.3\textwidth}
   \includegraphics[width=\textwidth]{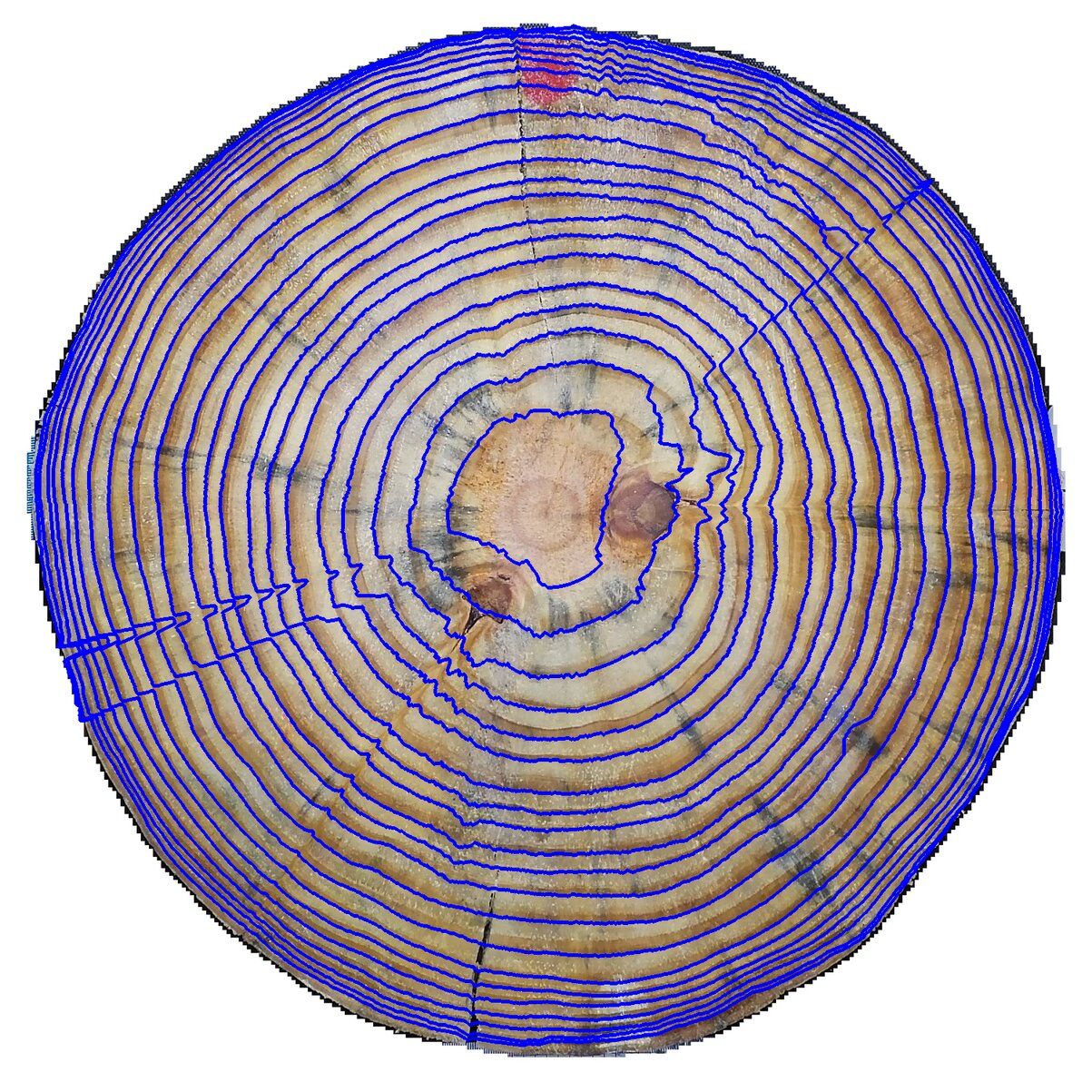}
   \end{subfigure}
    \begin{subfigure}{0.3\textwidth}
   \includegraphics[width=\textwidth]{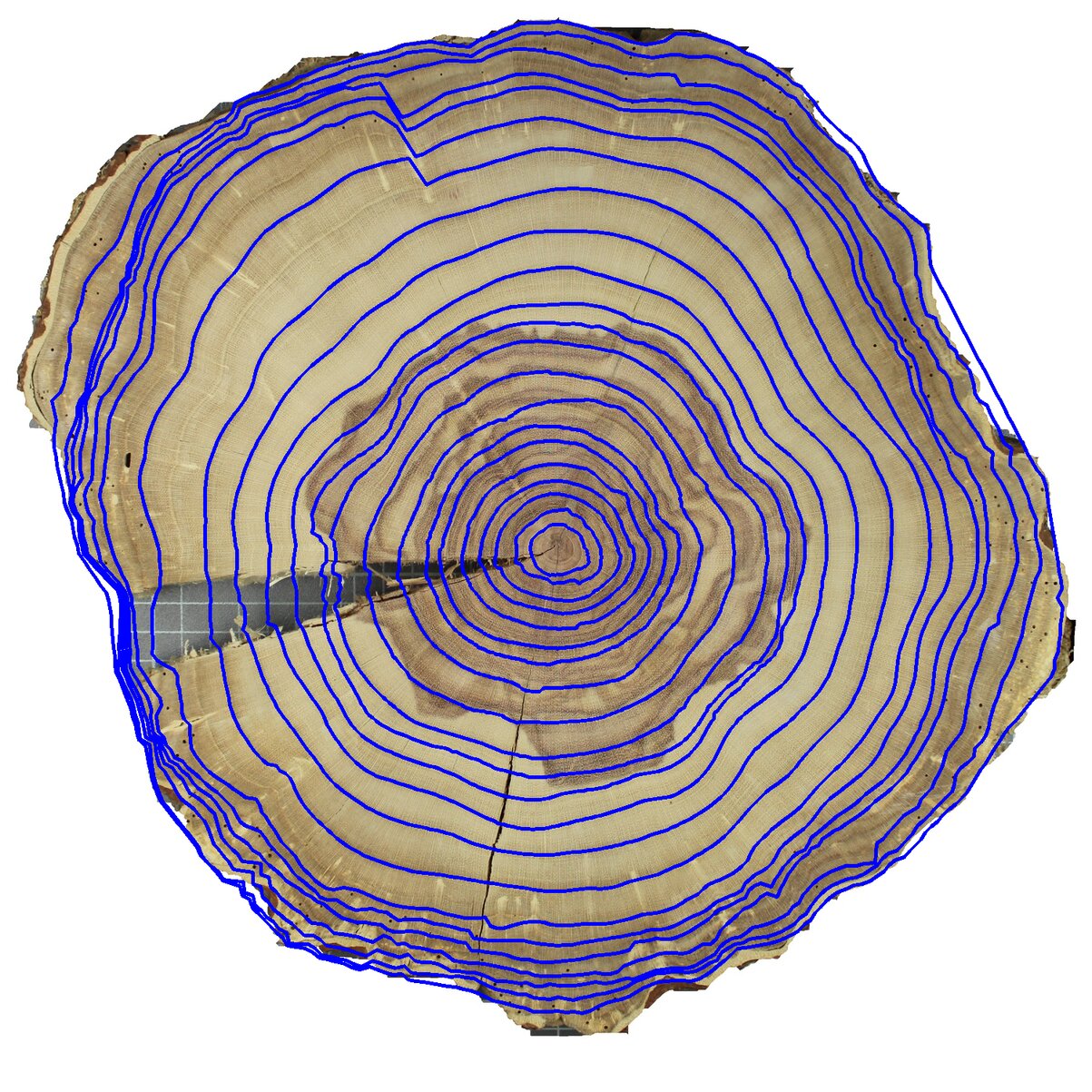}
   \end{subfigure}
   \begin{subfigure}{0.3\textwidth}
   \includegraphics[width=\textwidth]{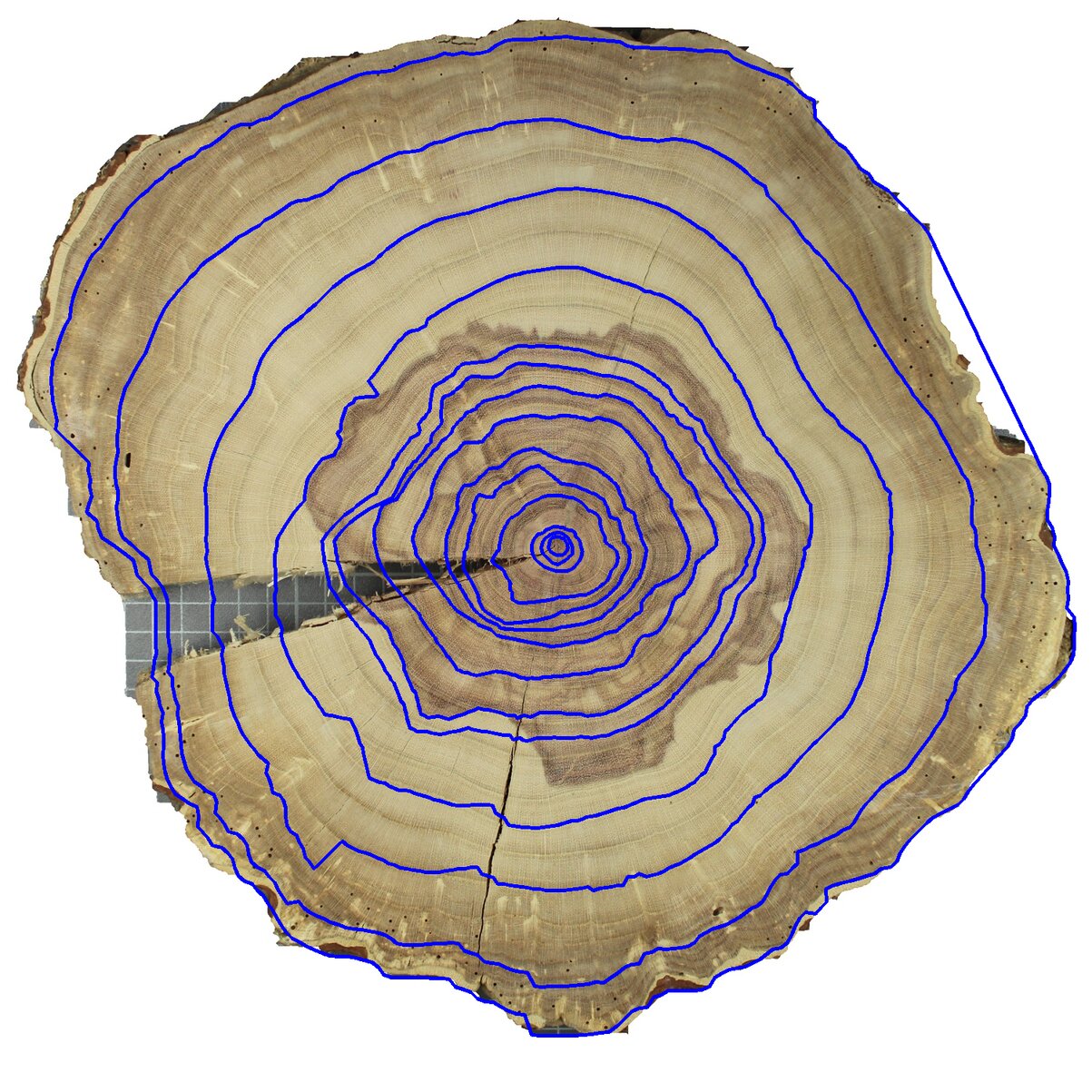}
   \end{subfigure}
   \begin{subfigure}{0.3\textwidth}
   \includegraphics[width=\textwidth]{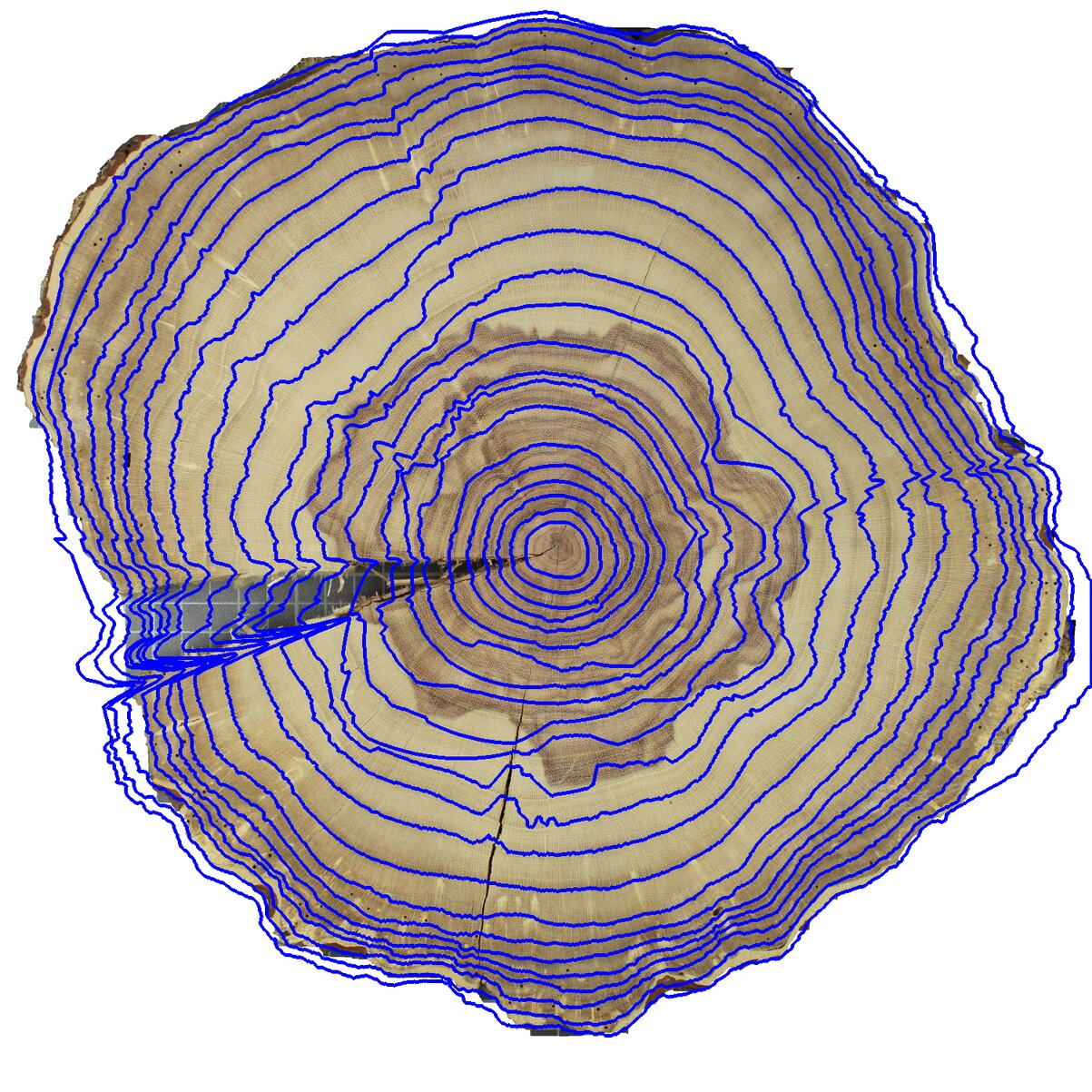}
   \end{subfigure}

    \begin{subfigure}{0.3\textwidth}
   \includegraphics[width=\textwidth]{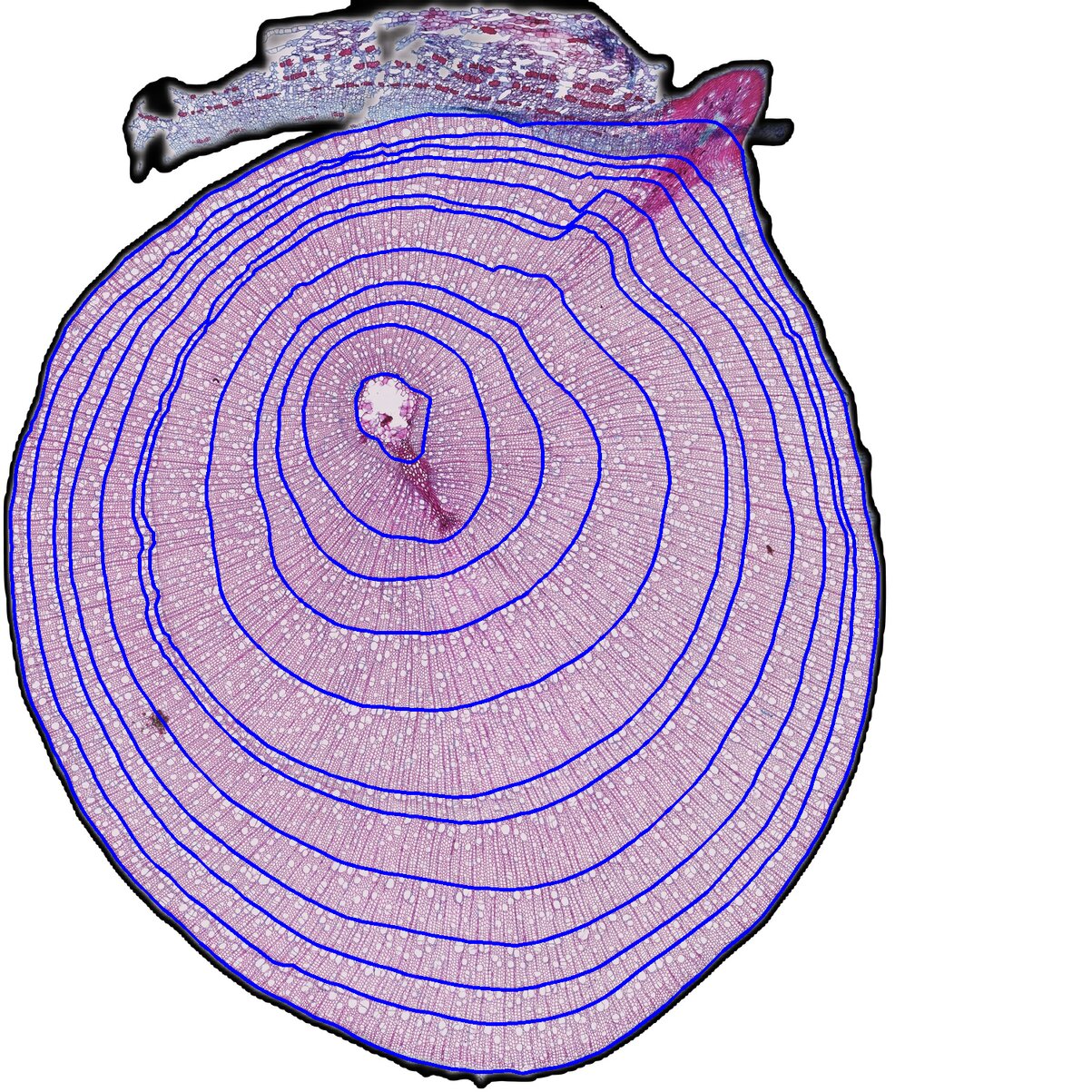}
   \caption{DeepCS-TRD}
   \end{subfigure}
   \begin{subfigure}{0.3\textwidth}
   \includegraphics[width=\textwidth]{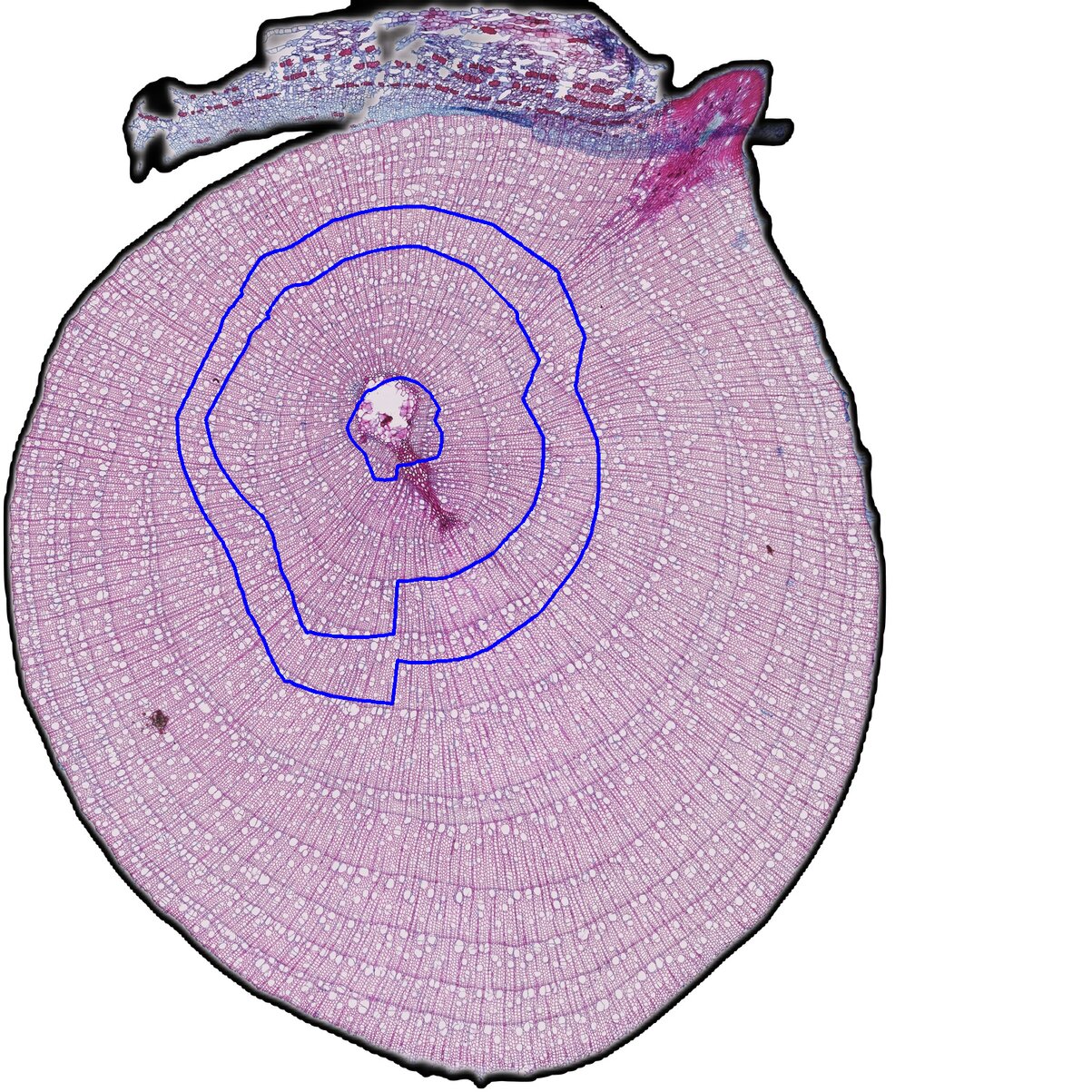}
   \caption{CS-TRD}
   \end{subfigure}
   \begin{subfigure}{0.3\textwidth}
   \includegraphics[width=\textwidth]{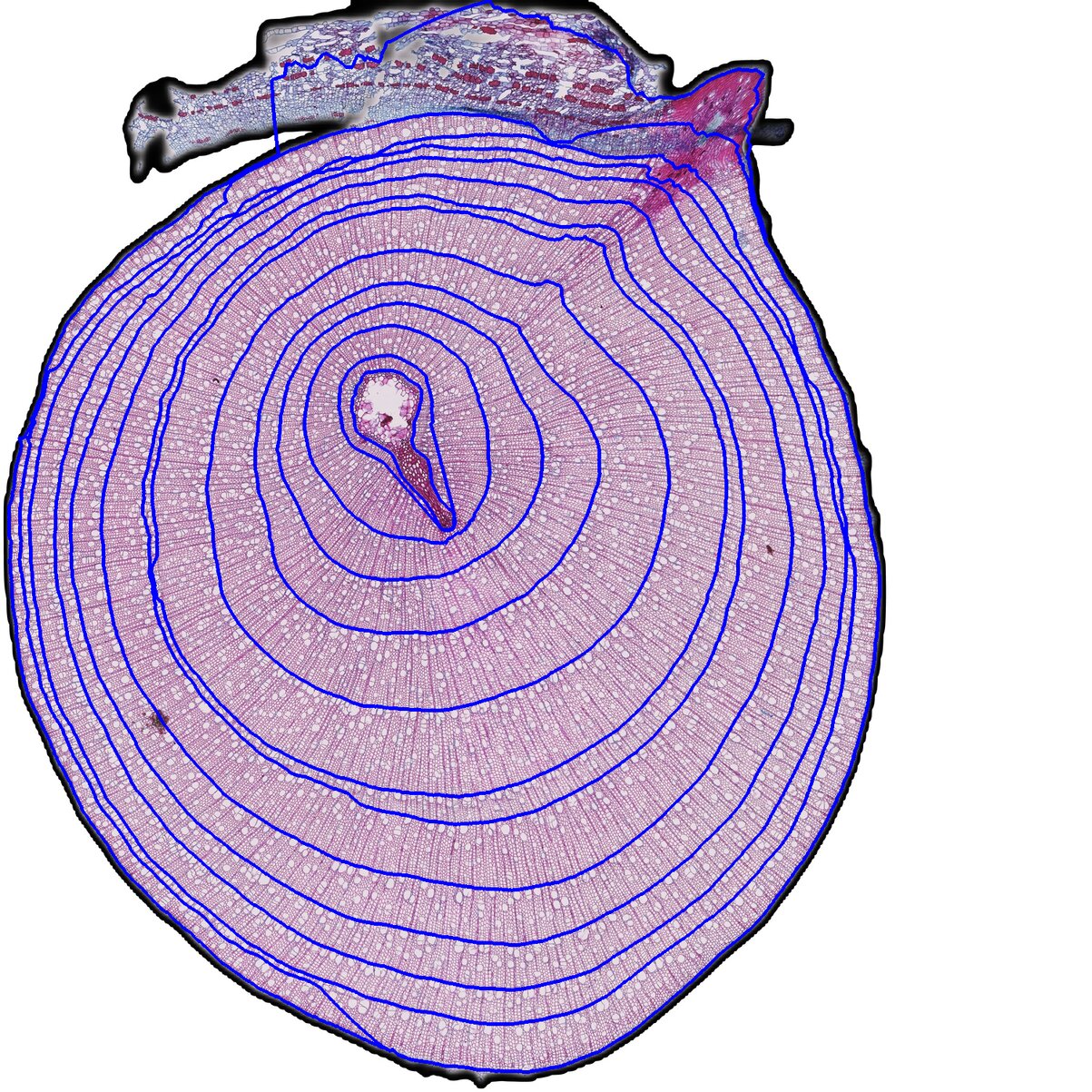}
   \caption{INBD}
   \end{subfigure}

   \caption{\textbf{Tree-Ring Delineation Results}. Each column displays the ring boundaries produced by each method, shown in blue. Each row corresponds to a different disc sample.  Column (a) DeepCS-TRD; Column (b) CS-TRD; Column (c) INBD. Note the presence of knots, cracks, fungus, and the differences between species. Note that the darker region in the third row does not correspond to the tree rings of that species.}
   \label{fig:resuls}
\end{center}
\end{figure}
\subsection{Results}

This section evaluates the performance of INBD, CS-TRD, and DeepCS-TRD on the test set across all datasets. The INBD method was modified to accept the pith boundary ground truth as input to ensure a fair comparison, as both DeepCS-TRD and CS-TRD take the pith locations as input but modeled as a pixel \cite{apd}.


\Cref{tab:results} presents each method's mAR and ARAND values across all datasets. The proposed DeepCS-TRD method has the best results on all the UruDendro datasets. It achieves outstanding results with near-perfect ring detection from the pith to the bark in UruDendro1 and UruDendro2, as illustrated in \Cref{fig:resuls}. Despite the INBD method giving very good performance in most of the samples, in some situations, it generates notorious propagation errors in the presence of significant fungal growth, cracks, or knots because it begins delineating ring boundaries from the pith center (see rows 1 and 2). In row 1's sample, DeepCS-TRD detects some false rings near the center, but its remaining detections are highly accurate. In contrast, CS-TRD gives good results, but rings are inaccurate with this strong fungus presence. In the sample in row 2, both DeepCS-TRD and CS-TRD performed well.

In the UruDendro3a dataset, DeepCS-TRD performs well; in the sample shown in \Cref{fig:resuls}, row 3, it correctly delineates 15 out of 19 rings. In contrast, the INBD method detects only six rings. CS-TRD detects five rings but interestingly identifies the three innermost rings missed by the other methods (see the region around the pith in row 3 for all methods).

In the DiskoIsland dataset, DeepCS-TRD performs slightly worse than the INBD method on average. In the sample in row 4, the INBD method correctly detects all the rings, while DeepCS-TRD missed the second ring. CS-TRD performs poorly in this dataset.

There are differences regarding the training time for the DeepCS-TRD and INBD methods. Both methods' maximum required training time was in the UruDendro1 dataset, with 25 and 4 hours for the INBD and DeepCS-TRD, respectively, using the same HW.  

\section{Conclusions}

A new method for automatic Tree-Ring delineation has been proposed. It adapts the CS-TRD method by replacing the edge detector step with a U-Net deep convolutional network. This modification allows us to apply the \textit{spider web} model to other species, such as \textit{Gleditsia triachantos} or \textit{Salix glauca}, with different, and often more difficult, annual ring patterns than for the coniferous species.  Additionally, the proposed method also increases the performance of CS-TRD for \textit{Pinus taeda} as can be seen in \Cref{tab:results}; in this case, the accuracy (mAR) has improved from 0.787 to 0.884 in UruDendro1 and from 0.710 to 0.809 in UruDendro2. 

By applying data augmentation techniques and the overlap-tile strategy, a high ring detection rate has been achieved in UruDendro3a despite using a limited number of images. Additionally, satisfactory results have been obtained on the \textit{Salix glauca} dataset. One advantage of this method over INBD is that INBD requires two training steps, whereas DeepCS-TRD requires one. Moreover, DeepCS-TRD significantly reduces training time, requiring just 4 hours compared to 25 hours for INBD on the UruDendro1 dataset.


\section{Acknowledgments}
The experiments presented in this paper used ClusterUY \cite{clusterUy}  (site: https://cluster.uy). This work was supported by project ANII\-FMV\-176061. CCP was supported by IRFD (702700133B) and H2020 CHARTER (869471).
{\small
\bibliographystyle{siam}
\bibliography{deepCSTRD}
}
\end{document}